%% file: main.tex
\pdfoutput=1
\documentclass[10pt,twocolumn,letterpaper]{article}

\usepackage{iccv}
\usepackage{times}
\usepackage{epsfig}
\usepackage{graphicx}
\usepackage{amsmath}
\usepackage{amssymb}
\usepackage{subcaption}
\usepackage{makecell}

\usepackage[breaklinks=true,bookmarks=false]{hyperref}

\iccvfinalcopy 

\def\iccvPaperID{****} 
\def\httilde{\mbox{\tt\raisebox{-.5ex}{\symbol{126}}}}

\ificcvfinal\fi
\begin{document}

\title{Learning Similarity Conditions Without Explicit Supervision}

\author{Reuben Tan\\
Boston University\\
{\tt\small rxtan@bu.edu}
\and
Mariya I. Vasileva\\
University of Illinois \\ at Urbana-Champaign\\
{\tt\small mvasile2@illinois.edu}
\and
Kate Saenko\\
Boston University\\
{\tt\small saenko@bu.edu}
\and
Bryan A. Plummer\\
Boston University\\
{\tt\small bplum@bu.edu}
}

\maketitle
\thispagestyle{empty}

\begin{abstract}
   Many real-world tasks require models to compare images along multiple similarity conditions (e.g. similarity in color, category or shape). Existing methods often reason about these complex similarity relationships by learning condition-aware embeddings. While such embeddings aid models in learning different notions of similarity, they also limit their capability to generalize to unseen categories since they require explicit labels at test time. To address this deficiency, we propose an approach that jointly learns representations for the different similarity conditions and their contributions as a latent variable without explicit supervision. Comprehensive experiments\footnote{\url{https://github.com/rxtan2/Learning-Similarity-Conditions}} across three datasets, Polyvore-Outfits, Maryland-Polyvore and UT-Zappos50k, demonstrate the effectiveness of our approach: our model outperforms the state-of-the-art methods, even those that are strongly supervised with pre-defined similarity conditions, on fill-in-the-blank, outfit compatibility prediction and triplet prediction tasks.  Finally, we show that our model learns different visually-relevant semantic sub-spaces that allow it to generalize well to unseen categories.
\end{abstract}

\section{Introduction}
\begin{figure}
    \centering
    \includegraphics[width=0.5\textwidth,trim=0cm 0.30cm 0cm 0cm, clip]{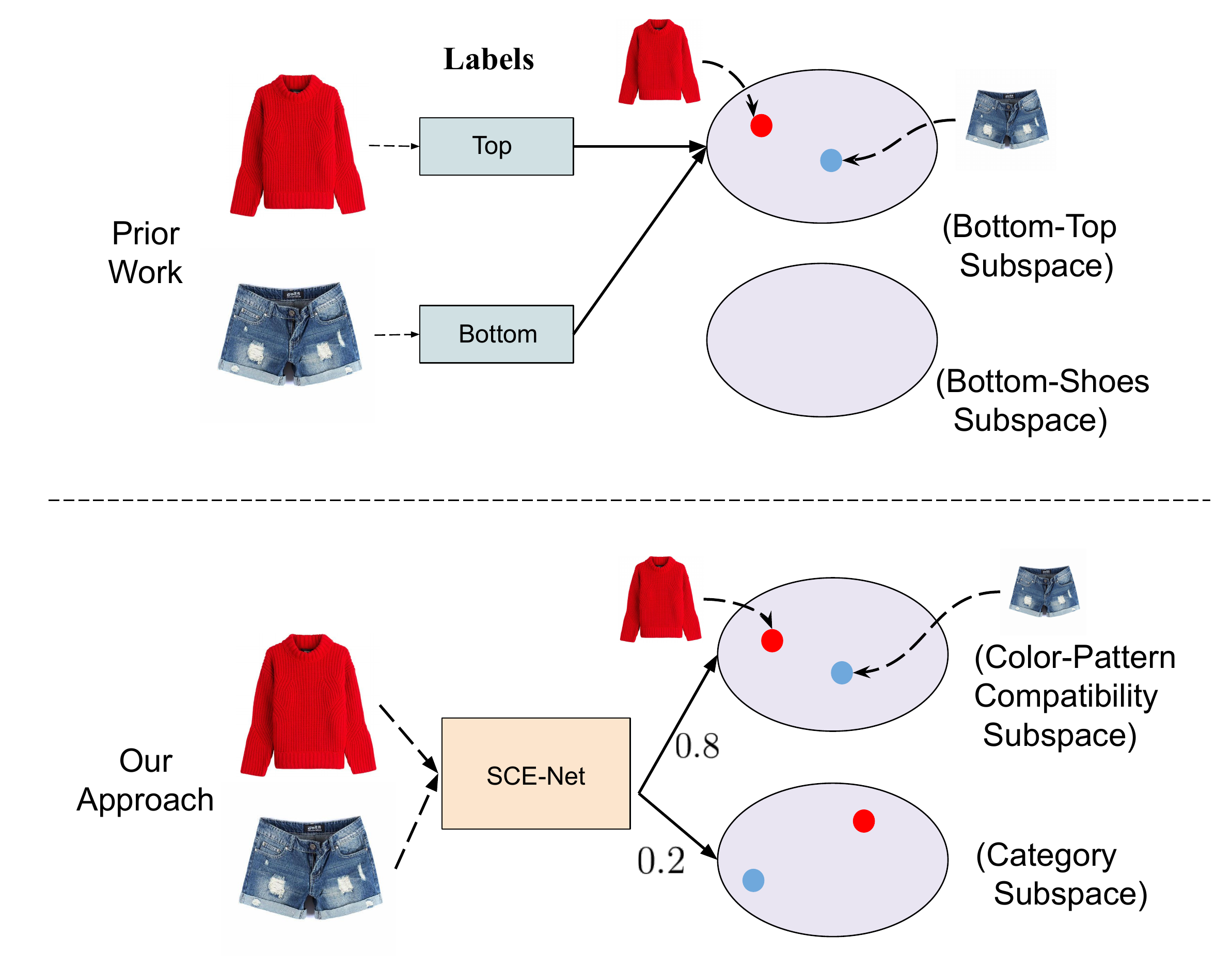}
    \caption{We propose the \emph{SCE-Net} model for learning multi-faceted similarity between images, such as compatibility of two fashion items. Previous work needed user-defined labels to learn multiple feature subspaces for measuring different aspects of similarity, \eg, one for comparing tops and bottoms and another for comparing bottoms and shoes (\eg, \cite{veit2017conditional,plummerHint2019,vasileva2018learning}). In contrast, our approach learns important subspaces without such labels in a data-driven manner.  The concepts and their contributions to a final similarity score are learned together as a single end-to-end trained model.
    }
    \label{fig:motiv}
    \vspace{-8pt}
\end{figure}


\begin{figure*}
\begin{center}
\includegraphics[width=\linewidth, height=6cm]{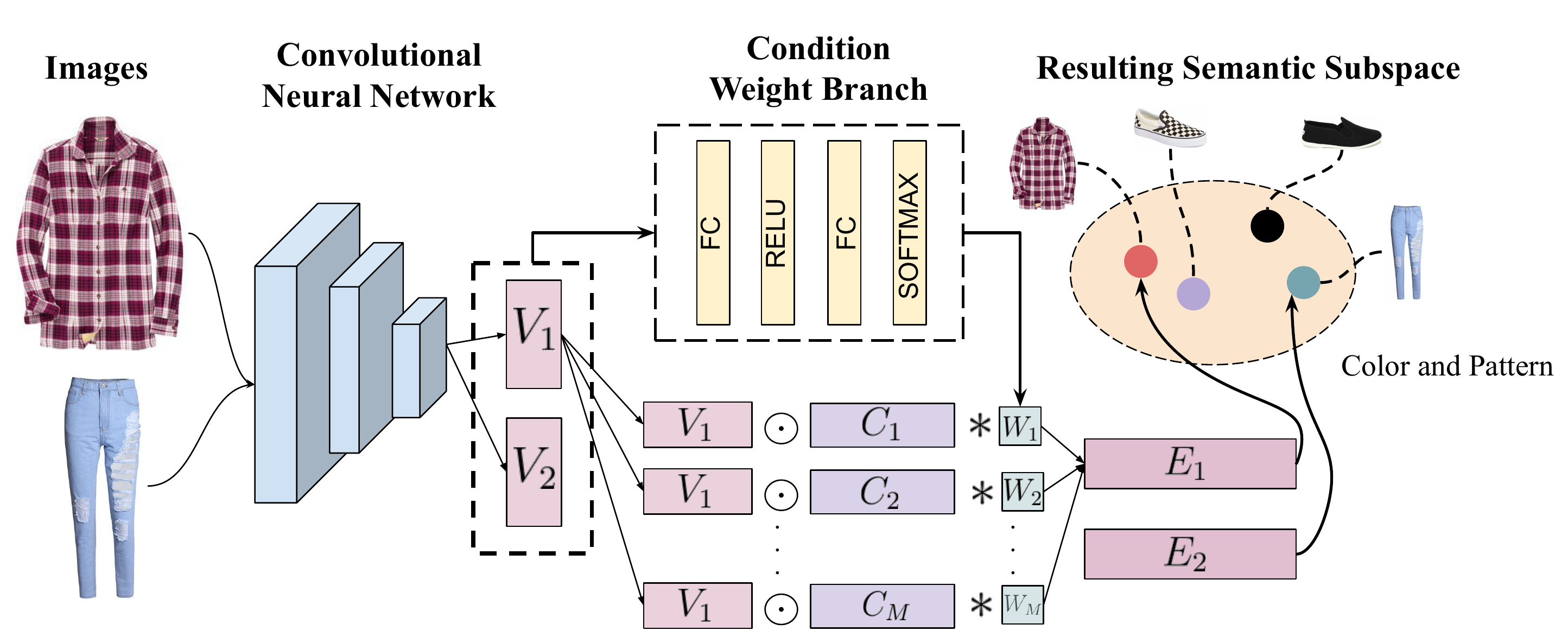}
\end{center}
   \caption{An overview of Similarity Condition Embedding Network (SCE-Net) which is trained end-to-end as a single model. The images are first passed into a Convolutional Neural Network (CNN) to extract their features, denoted $V_1$ and $V_2$, in a general embedding space.  To determine which semantic subspace is relevant to their comparison, both visual features are passed into the condition weight branch, which is a simple neural network. The output of the weight branch is a feature vector of dimension M, represented by $W_1, \cdot\cdot\cdot  ,W_M$.  It performs a dynamic assignment of the similarity condition masks, denoted $C_1, \cdot\cdot\cdot  ,C_M$, to the pair of images.  Each similarity condition mask $C_i$ has the same dimension D as the visual features and applied as a mask via element-wise product.  The masked embeddings are multiplied by the weight vector to produce the final representations $E_1$ and $E_2$.  These final representations induce a relevant semantic subspace within which the similarity between both images are compared.  We note that the subspace of 'color and pattern', shown in the figure, provides an example of possible notions of similarity encoded by the subspaces but we do not actually restrict the types of subspaces learned by the model.  The arrows from $V_2$ and to $E_2$ are removed to prevent the figure from being too crowded.}
\label{fig:model}
\vspace{-5pt}
\end{figure*}

Reasoning about the similarity between images or data of different modalities is an inherent challenge in computer vision.  Beyond its prevalence in fundamental problems such as image-sentence retrieval \cite{wang2019learning, vendrov}, cross-domain image-matching \cite{shrivastava2011data, huang2015cross}, attribution learning \cite{cheng2018learning, singh2016end} and visual categorization \cite{qian2015fine}, it also has an increasingly prominent role in computer vision problems in the fashion and retail domains like outfit style modeling \cite{hsiao2017learning}, fashion item retrieval and recommendation \cite{gu2018multi, liu2016deepfashion} and automatic capsule wardrobe generation \cite{hsiao2018creating}.  Metric learning (the task of learning a distance function between features based on supervised similar/dissimilar pairs) is a common approach used to tackle the above-mentioned problems and is often addressed by learning representations for objects in a unified embedding space, where the distances provide a measure of their similarity.
However, this is not naturally representative of the real world.  Objects can usually be described with multiple visual attributes such as color, shape or category.  Consider the example where a red shirt is similar to a pair of red shoes in color but dissimilar in object category. A single embedding space is unable to learn representations for these contradicting notions of similarity. By discounting such valuable information,  these embeddings are not able to reason comprehensively about the relative similarity between objects.  
There has been a recent trend of training embedding models conditioned on some given axis of similarity such as the object category (\eg, \cite{veit2017conditional,plummerHint2019,vasileva2018learning}) in order to learn disentangled representations (\ie illustrated on the top part of Figure~\ref{fig:motiv}).  This helps simplify complex similarity relationships by allowing the model to focus on only one similarity condition at a time for each semantic subspace.  However, by relying on such labels, these approaches cannot generalize to unseen categories and attributes, one of the primary advantages of embedding models.  As such, we seek to learn multiple notions of similarity jointly without explicit supervision via user-defined labels.

In this paper, we aim to learn how to separate the data, where the different similarity conditions and their contribution are treated as a latent variable and learned in a weakly supervised manner.  To obtain richer representations of visual similarity, we propose a Similarity Condition Embedding Network (SCE-Net) model that jointly learns multiple similarity conditions from a unified embedding space.   An illustrative overview of our model is provided in Figure \ref{fig:model}.  To begin, images are projected into a unified embedding space using a convolutional neural network.  The core component of our model is a set of parallel similarity condition masks, denoted as $C_{1}, \cdot\cdot\cdot, C_{M}$ in Figure \ref{fig:model}.   These masks are applied on the image features in the general embedding space.  By reweighting dimensions that are relevant to a specific notion of similarity, each similarity condition mask is encouraged to learn representations that encode different semantic subspaces.  The relevance of each condition mask to the objects is determined by a weight branch conditioned on their visual representations in the unified embedding space. The condition weight branch can be thought of as a type of attention mechanism \cite{xu2015show} that performs a dynamic assignment of each condition mask to the objects being compared.

Our work on learning disentangled representations is motivated by the Conditional Similarity Networks (CSN) of Veit~\etal \cite{veit2017conditional}.  The CSN model pre-defined similarity conditions to supervise the learning of disentangled representations. Our model attempts to learn such representations without explicit supervision via such pre-defined conditions. Plummer~\etal~\cite{plummerHint2019} found that considering the global similarity between items during training a CSN model produced more human-intuitive embedding spaces in addition to improving performance. Vasileva~\etal\cite{vasileva2018learning} adapted the CSN model to learn type-aware embeddings for modeling outfit compatibility.  Another drawback of these approaches is that they exhibit linear (\cite{veit2017conditional,plummerHint2019}) or quadratic~\cite{vasileva2018learning} growth in the number of conditions per desired similarity condition. In contrast, we found that we can often achieve better performance with far fewer learned subspaces (\eg, Vasileva~\etal learns 66 conditional subspaces for the fashion compatibility task on Polyvore Outfits whereas we obtain better performance with 5 learned subspaces). 

The contributions of our paper are summarized below:
\begin{itemize}
  \item We propose the Similarity Condition Embedding Network (SCE-Net), which learns richer representations of different notions of similarity from images without explicit category or attribute supervision.
  \item We demonstrate that SCE-Net generalizes well to novel categories and attributes in zero-shot tasks.
  \item Most importantly, we demonstrate that a dynamic weighting mechanism is integral in aiding a weakly supervised model to learn representations for different notions of similarity.
\end{itemize}

We perform extensive experiments over three datasets, Polyvore-Outfits \cite{vasileva2018learning}, Maryland-Polyvore \cite{han2017learning} and UT-Zappos50K \cite{yu2014fine}, where our approach outperforms the state-of-the-art in outfit compatibility prediction, fill-in-the-blank outfit completion and triplet prediction tasks, respectively, without requiring strong supervision (via category or attribute labels) used in prior work at test time.

\section{Related Work}
\textbf{Metric Learning.} Substantial prior work \cite{wang2014learning, chopra2005learning, hadsell2006dimensionality} has focused on measuring similarity between images in a single similarity context.  To achieve this, images are typically projected into a general embedding space where the respective distances between objects provide a measure of their relative similarity.  One notable shortcoming of this approach is that it does not consider different types of visual features.  In response to this, there has been a recent trend of comparing images across multiple axes of similarity.  As discussed in the introduction, several papers have proposed methods that attempt to learn disentangled representations which capture the different notions of similarity via supervision by pre-defined similarity conditions~\cite{veit2017conditional,plummer2018conditional,vasileva2018learning}. However, since these approaches are trained to only compare items along a known axis of similarity, they cannot make predictions between novel categories at test time.  Our idea of overcoming this restriction by using a similarity condition weight branch is similar to the phrase localization approach utilized by Plummer~\etal~\cite{plummer2018conditional}.  However, their work is primarily focused on measuring similarity between image regions and text, and their conditions are also supervised by text descriptions.  
Learning distance metrics has also attracted a lot of interest from the computer vision community.    Hsieh~\etal \cite{hsiao2017learning} utilizes collaborative filtering with implicit feedback to learn a joint metric which encodes user-user and user-item similarity while Sohn~\etal \cite{sohn2016improved} introduce a multi-class $N$-pair loss objective to improve deep metric learning.

\textbf{Visual Attributes.} Visual attributes (\eg color and pattern) entail a lot of information and have been shown to be an effective mode of communication between both humans and artificial agents \cite{das2017learning, batra2017cooperative}.  For example, Batra~\etal \cite{batra2017cooperative} seeks to improve the performance of agents by using visual attributes as their main mode of communication.  Attributes have also been used to address tasks such as image search and classification \cite{kumar2011describable, kovashka2012whittlesearch} and scene understanding \cite{shao2015deeply, patterson2012sun, li2010objects}.  However, one major limitation that researchers often face is the sparsity of supervision (\ie, a lack of example images and/or labels).  To address this, Yu~\etal \cite{yu2017semantic} trains attribute ranking models on synthetic images to determine the relevance of each attribute for the comparison of a pair of images. Others focus on ways of automatically discovering attributes in images \cite{berg2010automatic, rastegari2012attribute, vittayakorn2016automatic, ferrari2008learning}. For example, Ferrari~\etal \cite{ferrari2008learning} introduce a probabilistic generative model of visual attributes as well as an approach to learn its parameters from images. 


\textbf{Recommendation and Retrieval.} Similarity learning has also been used extensively to solve computer vision problems in other domains such as fashion and retail (\eg, \cite{han2017learning, vasileva2018learning, veit2015learning}). Using visual attributes is a naturally intuitive way to describe fashion items (\eg color, cut and style).  As such, identifying relevant attributes in visual representations of fashion items is essential to reasoning about similarity between them.  The deficiency of comparing images by projecting them into a general embedding space as described above is especially apparent in prior work on modeling fashion outfit compatibility \cite{li2017mining, han2017learning, vasileva2018learning, veit2015learning}.  In their approach, Veit~\etal \cite{veit2015learning} do not distinguish items by their types but instead attempt to learn the concepts of compatibility and  similarity  from  heterogeneous  dyadic  co-occurrences of items in user data. These visual attributes also form the basis of many interactive fashion search engines and recommendation systems \cite{zhao2017memory, al2017fashion, hsiao2018creating, plummerHint2019}.       
\section{Similarity Condition Embedding Network}
\label{sec:SCEnet_Model}

In this section we describe SCE-Net, our model which jointly learns representations for the different similarity conditions that may be present in a dataset by treating them and their contributions as a latent variable.  This allows us to train our end-to-end model in a weakly supervised manner where we only know if a pair of images are similar under some unknown condition. To begin, the images are projected via a CNN into a common feature space which we term as the general embedding space.  We denote this operation as $g(\textbf{x}; \theta)$ where \textbf{x} and $\theta$ represent the sets of images and parameters respectively. Our network consists of two components - a set of parallel similarity condition masks which we will discuss in Section~\ref{subsec:sce}, and a condition weight branch we will discuss in Section~\ref{subsec:condition_weight_branch}.  We discuss variants of our condition weight branch with inputs of different modalities in Section~\ref{sec:mm_scenet}.    

\subsection{Learning Similarity Conditions}
\label{subsec:sce}
A core component of our model is a set of $M$ parallel similarity condition masks of dimension $D$, denoted as $C_1, \cdot\cdot\cdot , C_M$ in Figure \ref{fig:model}. The value of $M$ is determined experimentally using held out data.  The similarity condition masks are applied, via elementwise product, to the image features in the general embedding space and their bearing on a similarity relationship is learned during training. By re-weighting relevant dimensions, the similarity condition masks are projecting the image features into secondary semantic subspaces of $\mathbb{R}^D$ which encode different similarity substructures.  For each similarity condition mask $C_j$ and general image feature $V_i$ , the masking operation is performed as follows:
\begin{equation}
E_{ij} = C_j \odot V_i,
\end{equation}
where $E_{ij}$ is the masked embedding and $\odot$ denotes the Hadamard product.  The output of the masking operation over all similarity condition masks and an image feature $v_i$ is a matrix of dimensions $M \times D$.  Let $O$ represent the output of the masking operation where $O = [E_{i1}, \cdot \cdot \cdot, E_{iM}]$.  Then, the final representation for the image feature is computed as a matrix-vector multiplication operation:
\begin{equation}
E_i = wO^T,
\end{equation}
where $w$ is the weight vector of dimension $M$ computed by the condition weight branch described below.

\subsection{Condition Weight Branch}
\label{subsec:condition_weight_branch}

Instead of pre-defining a set of similarity conditions, we use a condition weight branch to allow the model to automatically determine what concepts to learn. The condition weight branch determines the relevance of each condition mask based on the pair of objects being compared.  For a pair of images $x_i$ and $x_j$, the input feature to the condition weight branch is computed as follows:
\begin{equation}
y = concat\{V_i, V_j\},
\end{equation}
where $concat\{...\}$ denotes the concatenation operation.  As seen in Figure \ref{fig:model}, after concatenating these image features they are fed into a series of fully-connected and ReLU layers.  A softmax is used on the final activations resulting in a vector $w$ of dimension $M$ that is used to determine the relevance of each similarity condition mask to the objects being compared.

A triplet loss is a naturally intuitive way to learn representations with complex similarity relationships. We define a triplet of objects as a set $\{x_i, x_j, x_k\}$ where $x_i$ is the reference object and $x_j$ and $x_k$ are positive and negative objects that have been determined by the oracle to be semantically similar and dissimilar to $x_i$ under some unobserved condition $c$, respectively. In the context of this work, an oracle is defined to be a general entity that has the ground truth measures of similarity between all objects under the set of all possible similarity conditions.  Usually, the oracle takes the form of crowd-sourced datasets that are annotated with human labels. The final triplet loss is then given as:
\begin{equation}
l_{triplet}(x_i, x_j, x_k) =  max\{0,\; d(E_i, E_j) - d(E_i, E_k) + \mu \},
\end{equation}
where $d(E_i,E_j)$ denotes the Euclidean distance between the representations of objects $x_i$ and $x_j$ and the margin $\mu$ is a hyper-parameter. The triplet loss requires that $d(E_i, E_j)$ is smaller than $d(E_i, E_k)$ by a margin $\mu$ where the final image representations $E$ are computed as described above.

As in Veit~\etal \cite{veit2017conditional}, we impose an $l_1$ loss on the similarity condition masks to encourage sparsity and disentanglement.  In addition, we regularize the learned image representations $g(\textbf{x}; \theta)$ with an $l_2$ penalty.   As such, the final objective function for our model is given by:
\begin{equation}
l_{final} = l_{triplet}(\textbf{x})\;  + \lambda_{1}\;l_1 + \;  + \lambda_{2}\;l_2,
\end{equation}
where $\lambda_{1}$ and $\lambda_{2}$ are scalar hyperparameters.

\subsection{Multimodal Variants of SCE-Net}
\label{sec:mm_scenet}
In addition to the vision-only version of the condition weight branch used in our network, we also experiment with variations which leverages multimodal features that may provide some semantic relationship between the different conditions we wish to learn.  These variants are:
\smallskip

\textbf{Text Features.} We use the word `text' to refer to both sentences which may represent either the category labels or natural language descriptions of the images.  A sentence is tokenized and each token is represented using a pre-trained word embedding (\eg, \cite{pennington2014glove, mikolov2013distributed}).  For a pair of text features $(T_i, T_j)$ corresponding to image pair $(x_i, x_j)$, the input feature to the condition weight branch is computed according to the formulation above:
\begin{equation}
y = concat\{T_i, T_j\}.
\end{equation}

\textbf{Visual-Text Features.} For a pair of image features $(V_i, V_j)$ and their text features $(T_i, T_j)$, the condition weight branch determines the relevance of each condition embedding based on the input feature:
\begin{equation}
y = concat\{(V_i \odot T_i), \;(V_j \odot T_j)\}.
\end{equation}
We note that there are different ways to combine visual and text features such as concatenation and projections of both modalities into the same embedding space but elementwise product performed best in our experiments.

\section{Experimental Analysis}
We evaluate the capability of the SCE-Net model to capture different notions of similarity as well as how well it generalizes to novel image categories that are not seen during the training process.  To provide a fair comparison~\footnote{Recently, \cite{cucurull2019context} proposed a fashion compatibility model and evaluated on the Maryland Polyvore dataset, but it was published after our submission and thus, should be considered concurrent work.  In addition, they use a larger base network, ResNet-50 (theirs) vs.\ ResNet-18 (ours); we omitted their results since they are not directly comparable.} of our approach to other baseline models, we perform experiments on the Maryland-Polyvore \cite{han2017learning}, Polyvore-Outfits \cite{vasileva2018learning} and UT-Zappos50k \cite{yu2014fine} datasets.  The Maryland Polyvore and Polyvore Outfits datasets contain two evaluation tasks - outfit compatibility prediction and fill-in-the-blank (FITB). For outfit compatibility prediction,  the task is to evaluate the compatibility of a set of fashion items in an outfit.  As in Han~\etal \cite{han2017learning}, performance on this task is evaluated with the area under a receiver operating characteristic curve (AUC).  In the FITB experiment, given a set of candidate items and a subset of items in an outfit, the task is to select the most compatible candidate.  The effectiveness of the model is evaluated based on the overall accuracy.  Although using a larger final embedding has shown to have performance benefits (\eg,~\cite{han2017learning,vasileva2018learning}), this comes at a higher computational cost at test time.  We compare methods with the same final embedding size for a fair comparison.  We also evaluate the ability of our model to identify different relative strengths of attributes using the task triplet prediction of \cite{veit2017conditional} on the UT-Zappos50k dataset. We note that the level of supervision indicated in Tables \ref{tab:polyvore_outfits} and \ref{tab:zappos} refers to the amount of supervision required by ours and baseline models during test time (i. e. the models know explicitly which axis of similarity to compare the objects on).

\subsection{Datasets}
\textbf{Maryland Polyvore~\cite{han2017learning}.} This dataset collected 21,799 outfits from the social commerce website Polyvore. We use the outfit splits provided by the authors consisting of 17,316 outfits in the training set, 3,076 in the test set and 1,407 in the validation set.  In the test set provided by the authors, negatives in both the compatibility prediction and FITB tasks are sampled at random without consideration for item compatibility or category (\ie they could replace a ``top'' in an outfit with ``sunglasses'').  As such, we evaluate our model on a much more challenging test set provided by Vasileva~\etal~\cite{vasileva2018learning}, where the item category is taken into account when sampling for negatives.

\textbf{Polyvore Outfits~\cite{vasileva2018learning}.} This dataset is much larger than Maryland Polyvore, containing 53,306 outfits for training, 10,000 for testing and 5000 for validation. It is also sourced from the Polyvore website, but unlike the Maryland Polyvore dataset, it contains annotations for fine-grained item types and provides a text description of items.  

\begin{table*}[h!]
  \centering
  \label{tab:LABEL}
  \begin{tabular}{|l|c|c|c|c|c|}
  \hline
  & Test-time & \multicolumn{2}{|c|}{Polyvore Outfits} & \multicolumn{2}{c|}{Maryland Polyvore}\\
  \cline{3-6}
    Method & Supervision & Compat AUC & FITB Acc & Compat AUC & FITB Acc\\
    \hline
    Siamese Net \cite{vasileva2018learning} & None & 0.81 & 52.9 & 0.85 & 54.4\\
    Type-Aware Embedding Network  \cite{vasileva2018learning}& Strong & 0.86 & 55.3 & 0.90 & 59.9\\
    SCE-Net & None & \textbf{0.91} & \textbf{61.6} & \textbf{0.90} & \textbf{60.8}\\
  \hline
  \end{tabular}
  \newline
  \caption{Comparison of different methods on the outfit compatibility prediction and fill-in-the-blank tasks over the test set for Maryland Polyvore and Polyvore Outfits datasets.}
  \label{tab:polyvore_outfits}
\end{table*}

\textbf{UT-Zappos50k~\cite{yu2014fine}.}  This dataset contains 50,000 images of shoes with meta-data labels for annotations.  We use the triplets provided by Veit~\etal \cite{veit2017conditional} which are sampled based on four similarity conditions - type of the shoes, gender of the shoes, height of the shoe heels and the closing mechanism of the shoes.  Veit~\etal generated 200k train, 20k validation and 40k test triplets for each characteristic.  When training SCE-Net, we combine all the triplets from each characteristic into a single training set.

\subsection{Implementation Details}
\textbf{Maryland Polyvore and Polyvore Outfits.} For fair comparison, we adopt the implementation as detailed in Vasileva~\etal~\cite{vasileva2018learning}.  We use an 18-layer deep residual network \cite{he2016deep} as a shared feature extractor that has been pretrained on ImageNet \cite{deng2009imagenet} and fine-tuned during training on this task.  The features in the unified embedding space have an embedding size of 64 dimensions.  To represent the text descriptions, we also use the HGLMM Fisher vectors \cite{klein2015associating} of word2vec \cite{mikolov2013distributed} which have been PCA reduced to 6000 dimensions.  Vaslieva~\etal also took advantage of additional regularizers on their general embedding space (\ie, the output of $g(\textbf{x}; \theta)$ discussed in Section~\ref{sec:SCEnet_Model}) which helped improve performance. 
These include:

\begin{itemize}
  \item \textbf{VSE}: Visual-semantic loss which requires that an image $x_i$ is embedded closer to its description $t_i$ as compared to the other two images within a triplet.
  \item \textbf{Sim}: A loss which encourages similar images to embed nearby each other (analogously, similar text descriptions should also embed nearby each other).
\end{itemize}

For our experiments on both of these datasets, we included the VSE and Sim losses into our objective function.
As such, our objective function becomes:
\begin{equation}
l_{final} = l_{triplet}(\textbf{x})\;  + \lambda_{1}\;l_1  \;  + \lambda_{2}\;l_2 + \lambda_{3}\;l_{VSE} + \lambda_{4}\;l_{Sim},
\end{equation}
where $\lambda_{3}$ and $\lambda_{4}$ are scalar hyperparameters.
We use the same settings as Vasileva~\etal for learning rates and hyper-parameters for loss functions.

\textbf{UT-Zappos50k Dataset.} An 18-layer ResNet is also used as our base image encoder on this dataset.  Due to the triplet format of the dataset, we modify the weight branch to be conditioned on all three images in a triplet.  Given a triplet $\{x_i, x_j, x_k\}$, the input to the condition weight branch (at both train and test time) is given as,
\begin{equation} \label{eq:12}
y = concat\{V_i, V_j, V_k\},
\end{equation}
where $V_i$, $V_j$ and $V_k$ are the representations of images $x_i$, $x_j$ and $x_k$ respectively.  
In Section~\ref{sec:zappos_results}, we demonstrate that conditioning the weight branch on triplet visual representations helps our model to learn the different notions of similarity explicitly defined in the dataset.

\subsection{Results}
\subsubsection{Polyvore Outfits and Maryland Polyvore}

\begin{figure*}
    \begin{subfigure}[b]{0.45 \textwidth}
\centering
        \includegraphics[width=\linewidth]{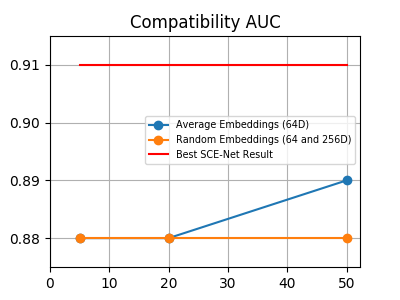}
        \caption{Compatibility AUC results}
        \label{fig:qual_seq}
\end{subfigure} \hfill
\begin{subfigure}[b]{0.45 \textwidth}
\centering
        \includegraphics[width=\linewidth]{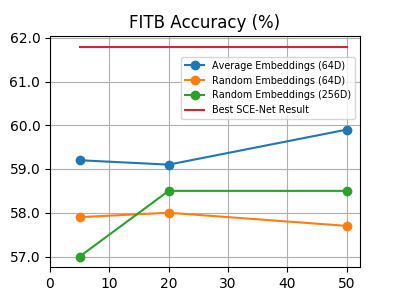}
        \caption{FITB Accuracy results}
        \label{fig:random_sample_frames}
\end{subfigure}
\caption{\small We report results of our model with average embeddings and random embeddings of 64 and 256D on the Polyvore-Outfit test set.  The values on the $x$-axis represent the number of similarity condition masks used in SCE-Net.  In both plots, the red line denotes the best result obtained by our SCE-Net model for comparison.}
\label{fig:plot}
\end{figure*}

Table~\ref{tab:polyvore_outfits} reports performance on the compatibility prediction and fill-in-the-blank tasks for the Maryland Polyvore and Polyvore Outfits datasets.  Across both datasets, our model obtains consistent improvements in both tasks over prior work.  In particular, our approach outperforms the state-of-the-art Type-Aware Embedding Network \cite{vasileva2018learning} by 5\% and 6.3\% on the compatibility prediction and FITB tasks respectively, demonstrating that it can better capture the compatibility relationship between items without requiring the type of each item being compared at test time.   In addition, we perform better using only 5 similarity conditions, whereas \cite{vasileva2018learning} learns 66 similarity conditions for Polyvore Outfits.

To show that our condition weight branch provides meaningful assignments, we compare to making random assignments of image pairs to conditions in Figure~\ref{fig:plot}.  We also compare to averaging the embeddings, which demonstrate that the additional parameters (from using multiple conditions for each image pair) in our approach cannot account for most of the improvements we see over \cite{vasileva2018learning} (which uses a single condition for each pair). The significant performance gap between SCE-Net and the average or random embeddings demonstrate that our dynamic weighing mechanism is integral to achieving good performance. We also show how the number of similarity conditions affect performance in Table~\ref{tab:ablation}, where we find that optimal performance can be obtained using only a few similarity conditions (\eg, 5 for Polyvore Outfits).


\begin{table}
  \centering
  \begin{tabular}{|c|c|c|}
  \hline
    Number of Conditions&Compat AUC&FITB Accuracy \\
    \hline
    1&0.86&53.2\\
    2&0.90&59.7 \\
    5&\textbf{0.92}&\textbf{62.1} \\
    10&0.91&60.8 \\
    20&0.89&59.7 \\
  \hline
  \end{tabular}
  \caption{Ablation studies on how the number of similarity condition masks effect the performance of our model on the validation set of  Polyvore Outfits.}
  \label{tab:ablation}
\end{table}

To evaluate the capability of our model to generalize to unseen categories based on visual features alone, we remove fashion items that belong to \emph{scarves} and \emph{accessories} categories from the training set. We selected these two categories because they are generally not an essential part of outfits and appear in fewer outfits than other categories in the training set.  For evaluation purposes, we extract FITB questions from the test set where the candidate choices belong to the removed categories.  As a baseline comparison, we train a Siamese network based off the model used by Vasileva~\etal on the modified training set.  The results for both models are reported in Table \ref{tab:unseen}.  Our model outperforms Siamese Net by a significant margin in both categories, demonstrating the ability of our model to generalize well to novel categories and attributes. 

\begin{table}
  \centering
  \begin{tabular}{|c|c|c|}
  \hline
  \multicolumn{3}{|c|}{Unseen Categories (FITB Accuracy)} \\
  \hline
    \thead{Method} & \thead{Scarves} & Accessories\\
    \hline
    Number of questions & 144 & 248 \\
    \hline
    Siamese Net & 46.62 & 50.82\\
    SCE-Net & \textbf{59.46} & \textbf{56.55}\\
  \hline
  \end{tabular}
  \caption{Comparison of different methods on a subset of FITB questions from the Polyvore-Outfits test set where the candidate choices belong to categories that are unseen during training.}
  \label{tab:unseen}
\end{table}

The performance of our multimodal variants are shown in Table~\ref{tab:variants}.  Surprisingly, using the language features of the items' labels alone leads to results that are comparable to those obtained by using the visual features of the items. Using a combination of visual and language features does not lead to a performance gain.  However, this could be due to that fact that the language features of item labels do not contain much semantic information.  It is possible that we can observe a larger improvement if the language features for the items' descriptions are used instead.  However, not all items in this dataset contain a corresponding description.

\begin{table}
  \centering
  \begin{tabular}{|c|c|c|}
  \hline
  \multicolumn{3}{|c|}{Variants of Condition Weight Branch} \\
  \hline
    Number of Conditions & Compat AUC & FITB Accuracy\\
    \hline
   Labels & 0.90 & 60.8\\
    Visuals & 0.91 & 61.6 \\
    Visual-Labels & 0.90 & 61.5 \\
  \hline
  \end{tabular},
  \caption{Results on the Polyvore-Outfit test set obtained by variants of the SCE-Net model with input features of different modalities into the condition weight branch.}
  \label{tab:variants}
\end{table}

\subsubsection{UT-Zappos50K}
\label{sec:zappos_results}

We evaluate the effectiveness of our approach on the task of triplet prediction against the strongly-supervised CSN model of Veit~\etal \cite{veit2017conditional}.  Recall that the test set is divided into 4 similarity conditions.  In particular, during inference, Veit~\etal evaluates each triplet with the query $\{x_i, x_j, x_k, c\}$ to determine if the distance between $x_i$ and $x_k$ is smaller than that of $x_i$ and $x_k$ under the similarity condition $c$.  Such explicit supervision during evaluation provides their model with an unfair advantage as compared to our proposed SCE-Net which isn't provided the similarity condition being compared.

Table~\ref{tab:zappos} shows that when using the concept weight branch to combine our weakly supervised conditions SCE-net outperforms the CSN model, which is provided the exact condition being compared, by approximately 3.2\% when using the same number of learned conditions (\ie, 4). Reducing the number of learned conditions by 1 for our model, we still outperform the CSN model by 2\%.  This suggests that it is beneficial to not limit the learning of a notion of similarity to a single subspace.  Instead, using a weighted combination of semantic subspaces encourages a model to learn better representations for a similarity condition. In addition, the number of similarity condition masks required for optimal learning increases with the number of similarity conditions present in the dataset.

\begin{table}
\setlength{\tabcolsep}{2pt}
  \centering
  \label{tab:LABEL}
  \begin{tabular}{|rl|c|c|}
  \hline
    & Method & Error Rate & \thead{Test-time \\Supervision}\\
    \hline \bf{(a)} &
    CSN fixed disjoint masks \cite{veit2017conditional} & 10.79\%  & Strong\\
    &CSN learned masks \cite{veit2017conditional} & 10.73\% & Strong\\
  \hline \bf{(b)} &
   SCE-Net (2) & 11.12\% & None\\
   & SCE-Net (3) & 8.48\% & None\\
   & SCE-Net (4) & \textbf{7.53\%} & None\\
  \hline
  \end{tabular}
  \caption{Results on the UT-Zappos50K test set. \textbf{(a)} contains the results reported in prior work \cite{veit2017conditional} and \textbf{(b)} reports the results of our model.  Numbers in parenthesis indicate the number of similarity condition masks used.}
  \label{tab:zappos}
\end{table}

\subsection{Visualizations of Learned Subspaces}
\begin{figure*}
    \begin{subfigure}[b]{0.45 \textwidth}
\centering
       \textbf{Similarity Condition 1}\par\medskip
        \includegraphics[width=\linewidth]{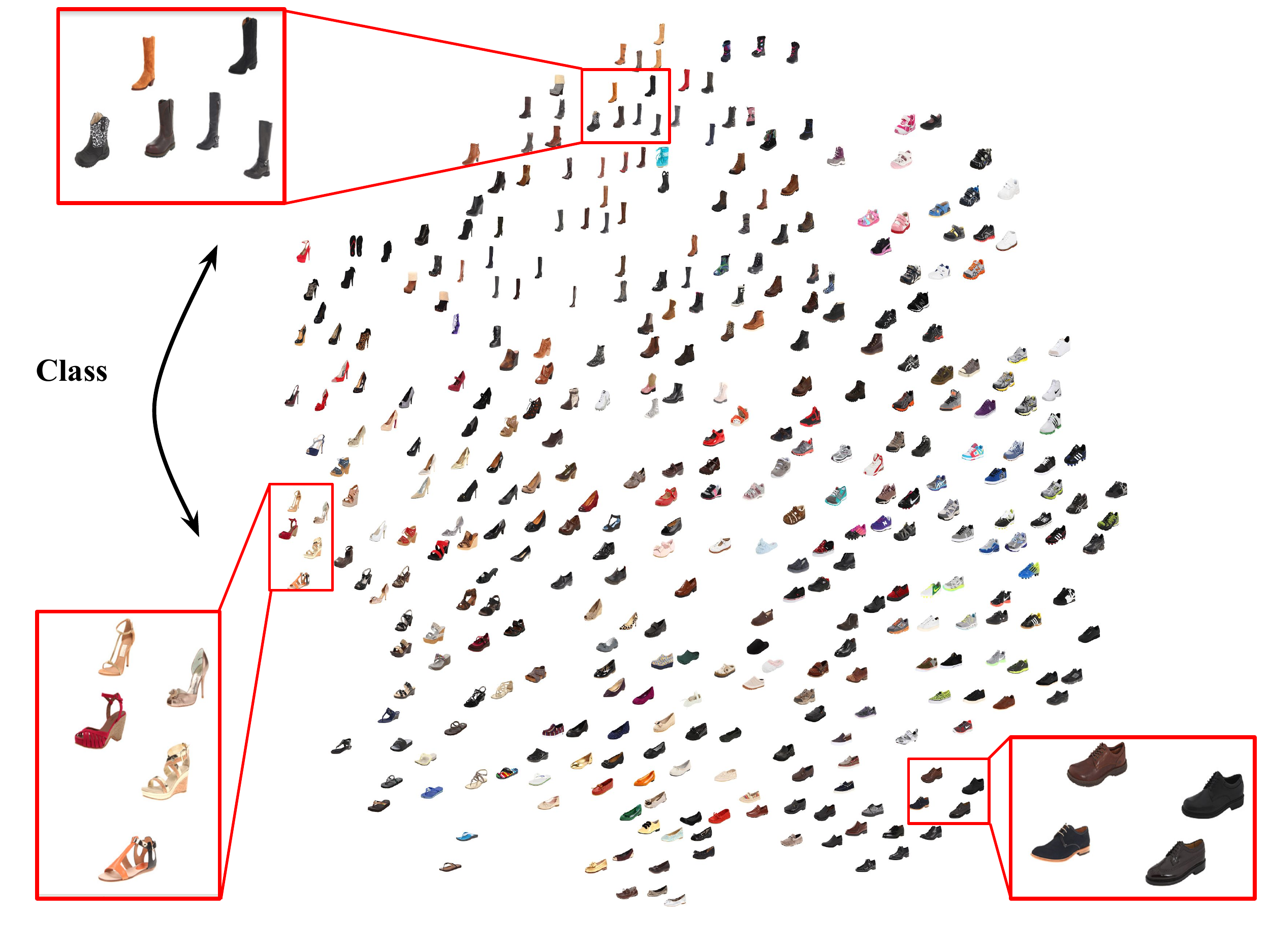}
        \caption{The visualization suggests that shoes are differentiated by class (\eg boots and slippers) in this subspace.}
        \vspace{32pt}
        \label{fig:vis_a}
\end{subfigure} \hfill
\begin{subfigure}[b]{0.45 \textwidth}
\centering
        \textbf{Similarity Condition 2}\par\medskip
        \includegraphics[width=\linewidth]{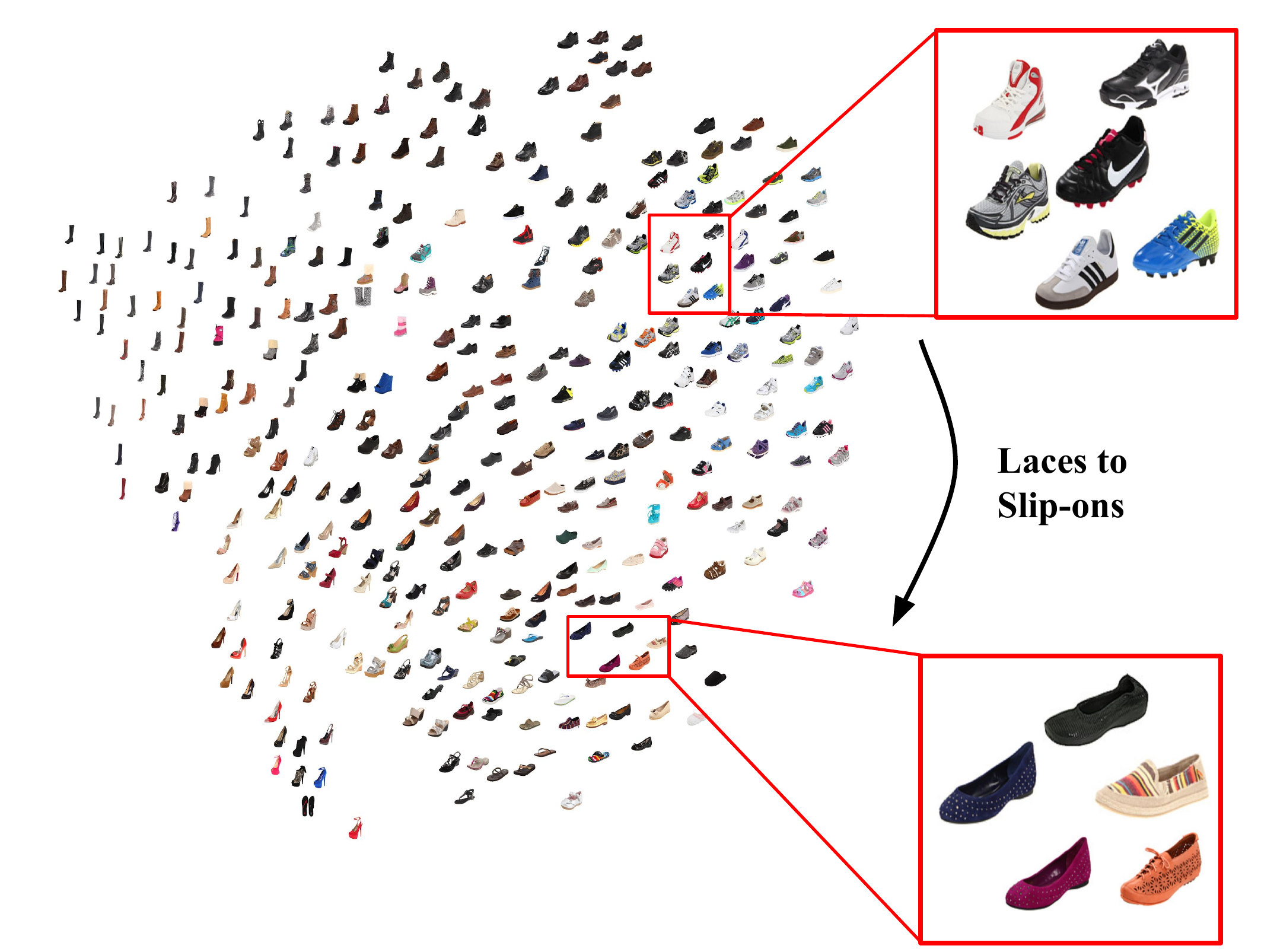}
        \caption{Shoes at the top of this subspace generally have laces while the shoes at the bottom are generally slip-ons, demonstrating that this similarity condition has learned to differentiate between closing mechanisms.}
        \vspace{11pt}
        \label{fig:vis_b}
\end{subfigure}
\begin{subfigure}[b]{0.45 \textwidth}
\centering
        \textbf{Similarity Condition 3}\par\medskip
        \vspace{-7pt}
        \includegraphics[width=\linewidth]{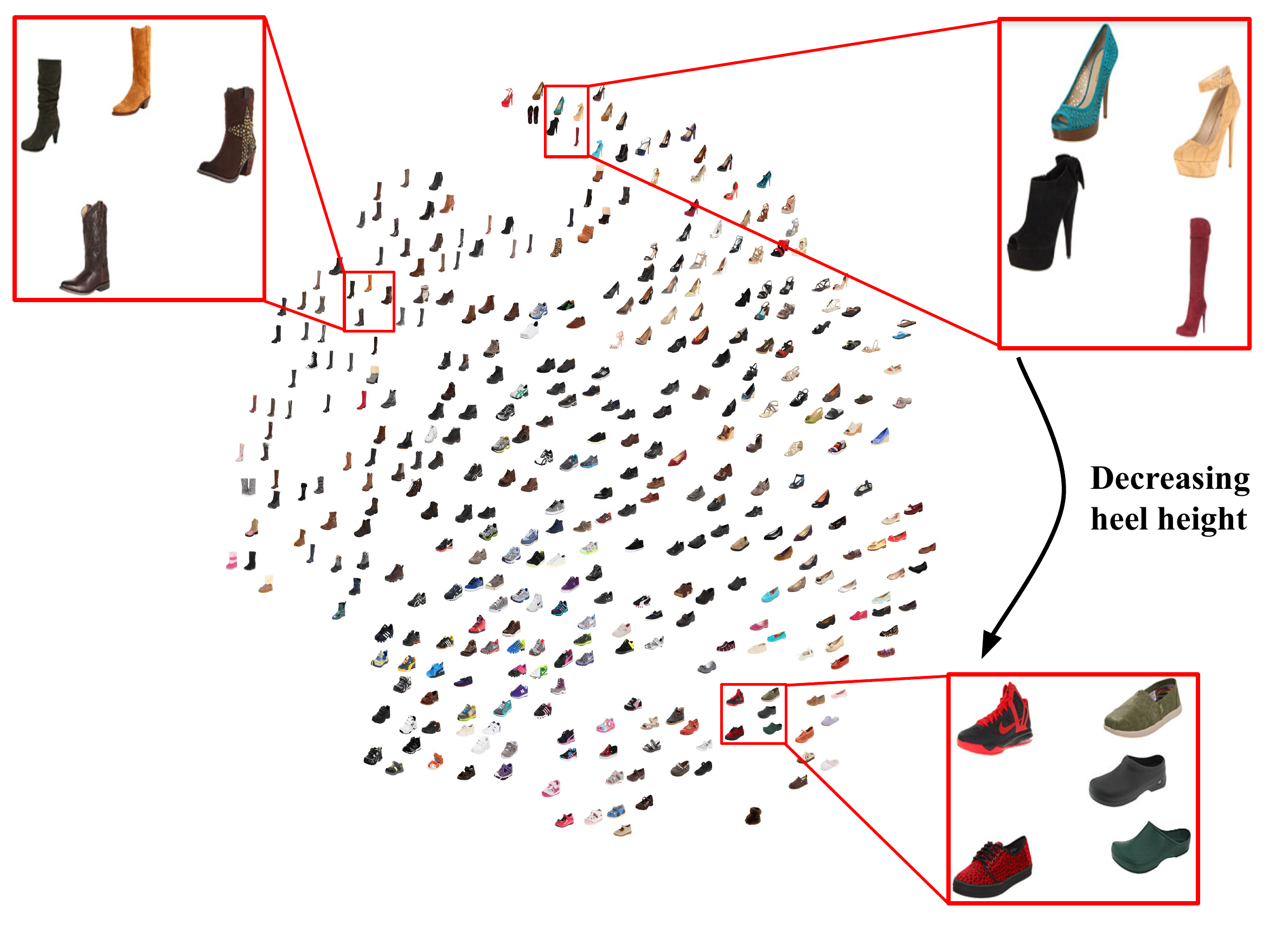}
        \caption{The visualization suggests that shoes are differentiated by heel height in this subspace. The heel height of the shoes decreases as we go from the top to the bottom of the subspace.}
        \label{fig:vis_c}
\end{subfigure} \hfill
\begin{subfigure}[b]{0.45 \textwidth}
\centering
        \textbf{Similarity Condition 4}\par\medskip
        \includegraphics[width=\linewidth]{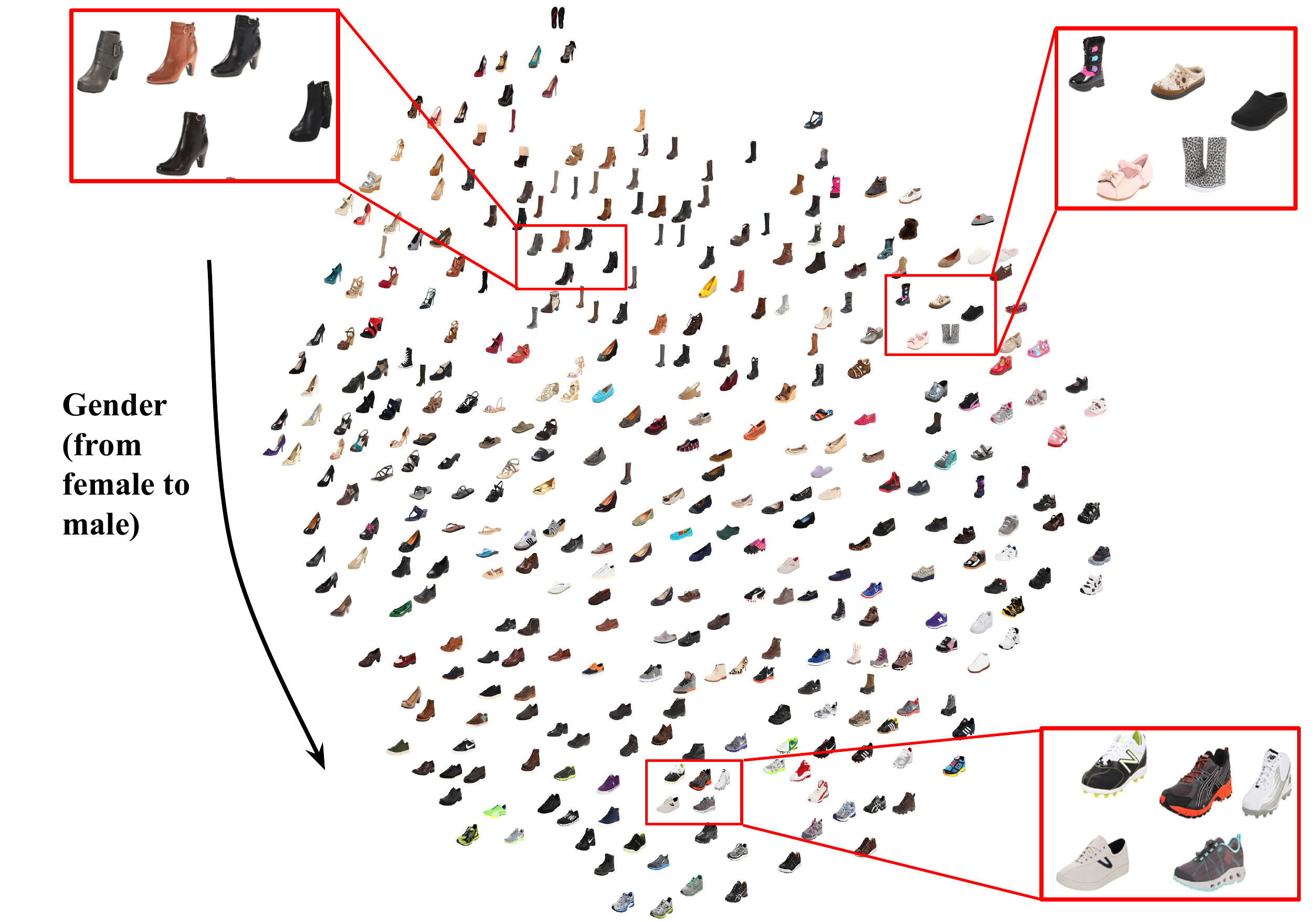}
        \caption{The visualization suggests that shoes are differentiated by gender in this subspace. Women's shoes are embedded at the top of the subspace, and men's shoes at the bottom.}
        \label{fig:vis_d}
\end{subfigure}
\vspace{-2pt}
\caption{\small Visualizations of the semantic subspaces encoded by our 4 similarity condition masks on the UT-Zappos50k dataset. }
\vspace{-5pt}
\label{fig:visualize}
\end{figure*}

To gain insights into the conditions learned by our model, we provide t-SNE \cite{maaten2008visualizing} visualizations for all similarity condition masks of learned subspaces for the UT-Zappos50k dataset in  Figure \ref{fig:visualize}.  The first similarity condition mask learns to differentiate shoes based on their class (\eg boots and high-heels). As we move from the top of the visualization in Figure \ref{fig:vis_b} to the bottom, we can clearly see that the closing mechanism of the shoes gradually changes from laces to slip-ons. Figure \ref{fig:vis_c} displays a subspace that learns the differences in the heel height.  In this case, the heel height of the shoes is decreasing from the top of the embedding space to the bottom.  From Figure \ref{fig:vis_d}, we see the fourth similarity condition mask has learned to differentiate shoes based on the targeted gender. Women's shoes are embedded at the top of the subspace while men's shoes are mostly embedded at the bottom.  This demonstrates that even with just weak supervision during training time, our approach is capable of learning visually-relevant similarity conditions that are explicitly defined in the dataset.

\section{Conclusion} 
In this work, we propose an approach that treats the different similarity conditions and their contributions as a latent variable and attempts to learn them in a weakly supervised manner.  SCE-Net removes the need for strong supervision via pre-defined similarity conditions by using a condition weight branch conditioned on visual representations of images to determine the context relevance of each similarity condition mask.  We demonstrate that our model not only outperforms strongly supervised methods but also generalizes well to novel image categories and attributes.  

We show that a dynamic weighting mechanism is essential in training a weakly supervised model to learn different notions of similarity. In particular, our results indicate that restricting the learning of a similarity condition to a single subspace can be disadvantageous to the learning capability of the model.  Finally, we demonstrate that a weighted combination of semantic subspaces can learn better representations for a similarity condition.  One exciting avenue for future work is to learn to determine the optimal number of similarity condition masks in an unsupervised manner.

\noindent\textbf{Acknowledgements:} This work is supported in part by DARPA and NSF awards IIS-1724237, CNS-1629700, CCF-1723379.

{\small
\bibliographystyle{ieee_fullname}
\bibliography{main}
}
\clearpage
\onecolumn
\input{appendix.tex}

\end{document}

%% file: appendix.tex




\iccvfinalcopy 

\def\iccvPaperID{3618} 
\def\httilde{\mbox{\tt\raisebox{-.5ex}{\symbol{126}}}}

\ificcvfinal\pagestyle{empty}\fi



\appendix
\maketitle
\thispagestyle{empty}

\section{Sensitivity of SCE-Net To Noisy Triplets}
We perform a series of ablation experiments on the UT-Zappos50k dataset to evaluate the robustness of our model to noisy, random triplets. In Table \ref{tab:zappos-reduced}, we report results where we added noise to our training triplets to measure how performance would be affected as the triplet quality degrades.  We find that our model is fairly robust to the inclusion of noisy triplets during training.  Replacing 12.5\% of training samples with random triplets on the UT-Zappos50k dataset did not seem to affect the performance of SCE-Net.  In fact, our approach only has a small increase in error rate even when 50\% of the training triplets were random triplets, getting an error rate of 10.04\%.  Notably, this result is comparable to the strongly supervised CSN [34] model trained without noise from random triplets.

\begin{table}[h]
  \vspace{-2pt}
  \centering
  \begin{tabular}{|l|c|c|c|}
  \hline
    Method&Number of & Percentage of & Error\\
    & Samples & Random Triplets & Rates\\
    \hline
    CSN [34]&200000&None&10.73\%\\
    SCE-Net&100000&None&11.37\%\\
    SCE-Net&200000&None&8.29\%\\
    SCE-Net&200000&12.5\%&7.44\%\\
    SCE-Net&200000&25\%&10.68\%\\
    SCE-Net&200000&50\%&10.04\%\\
  \hline
  \end{tabular}
  \caption{Ablation results on the quality and size of training triplets in the UT-Zappos50k test set.}
  \vspace{-10pt}
  \label{tab:zappos-reduced}
\end{table}


\section{Visualizations of Compatibility Relationships}
\begin{figure*}[!ht]
\begin{center}
\includegraphics[width=\linewidth]{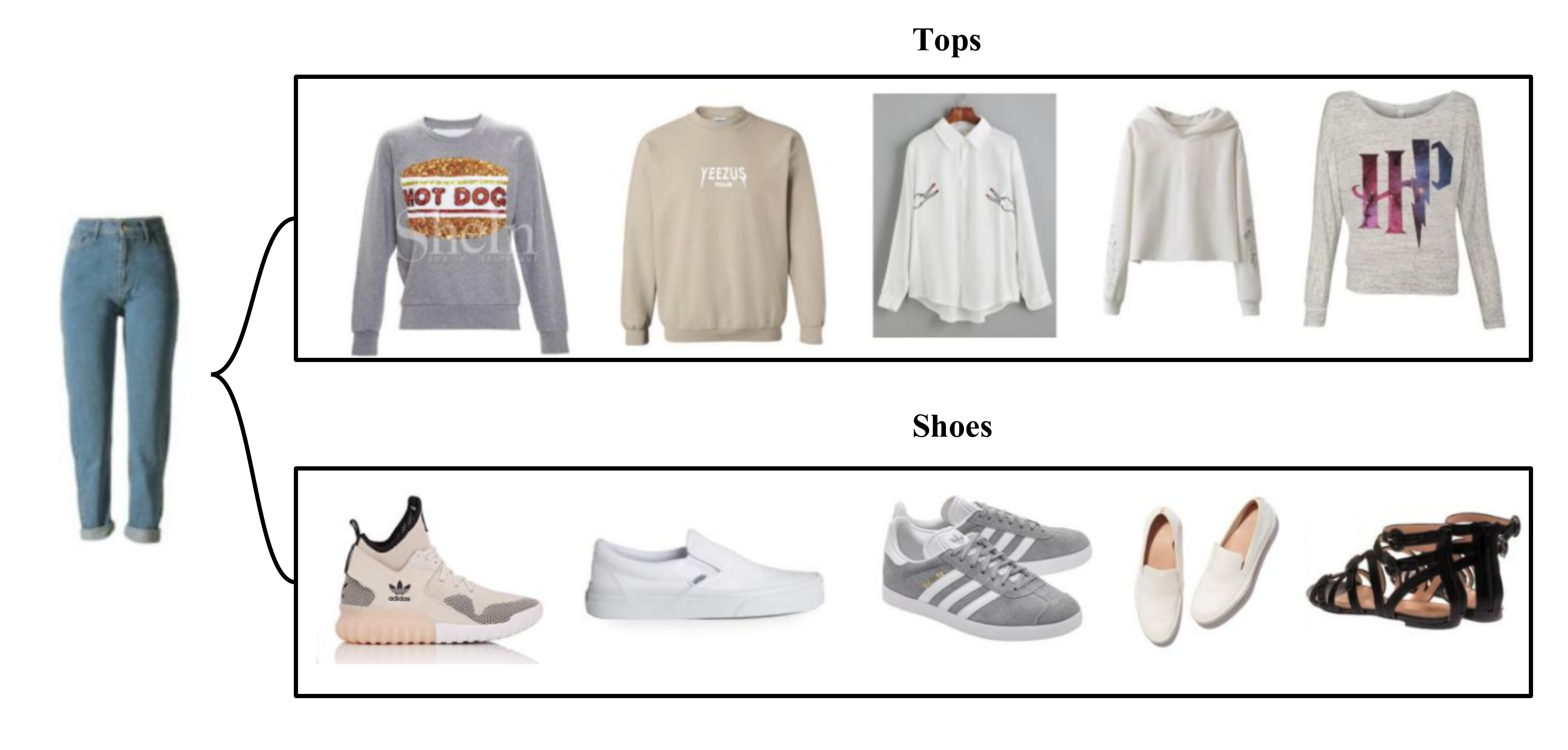}
\end{center}
\caption{The query item is a bottom and the items contained in the boxes are the top 5 most compatible tops and shoes.}
\label{fig:top-bottom-shoes-2}
\end{figure*}

\begin{figure*}[!ht]
\begin{center}
\includegraphics[width=\linewidth]{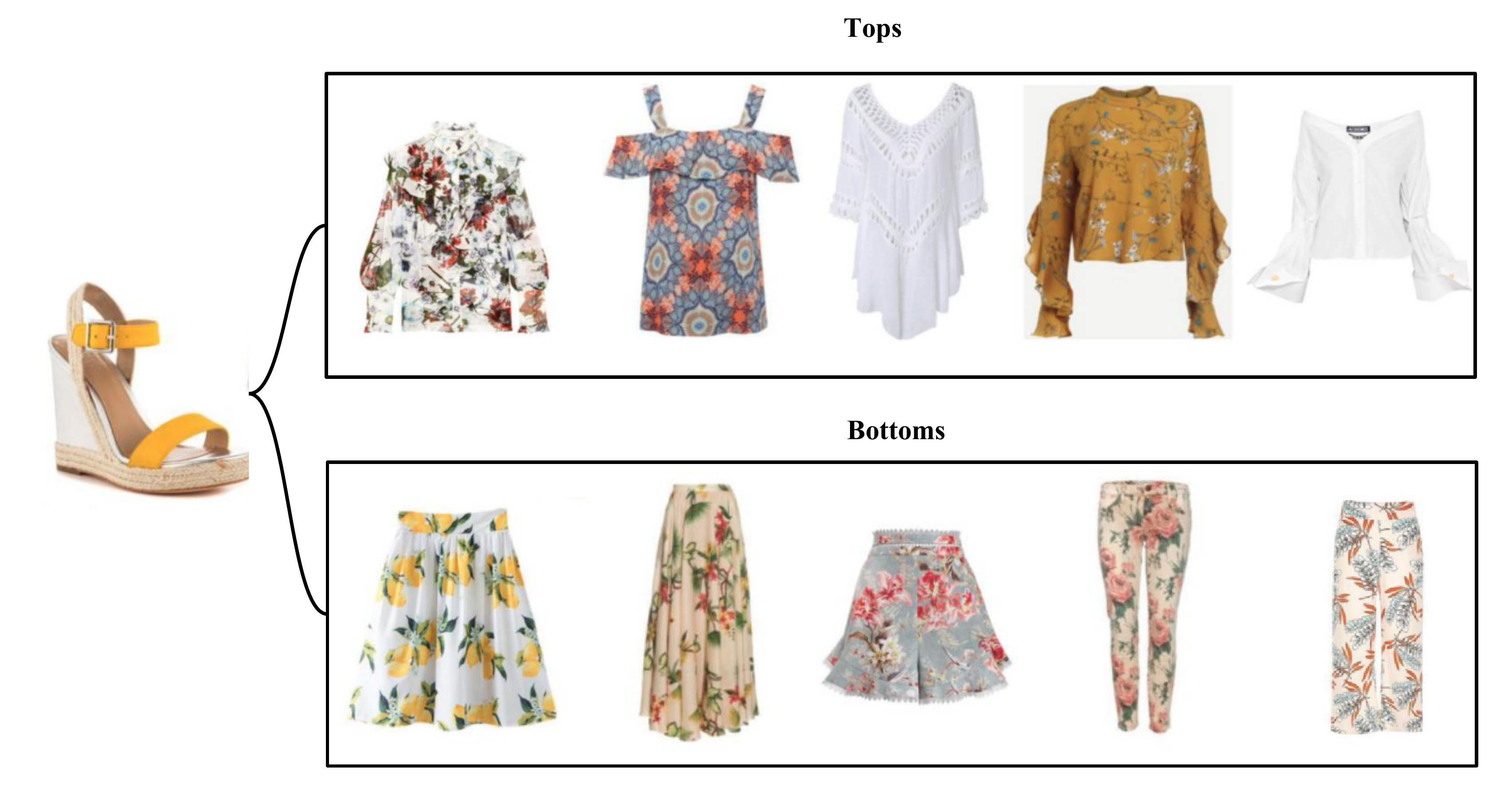}
\end{center}
\caption{The query item is a shoe and the items contained in the boxes are the top 5 most compatible bottoms and tops.}
\label{fig:top-bottom-shoes-3}
\end{figure*}

\begin{figure*}[!ht]
\begin{center}
\includegraphics[width=\linewidth]{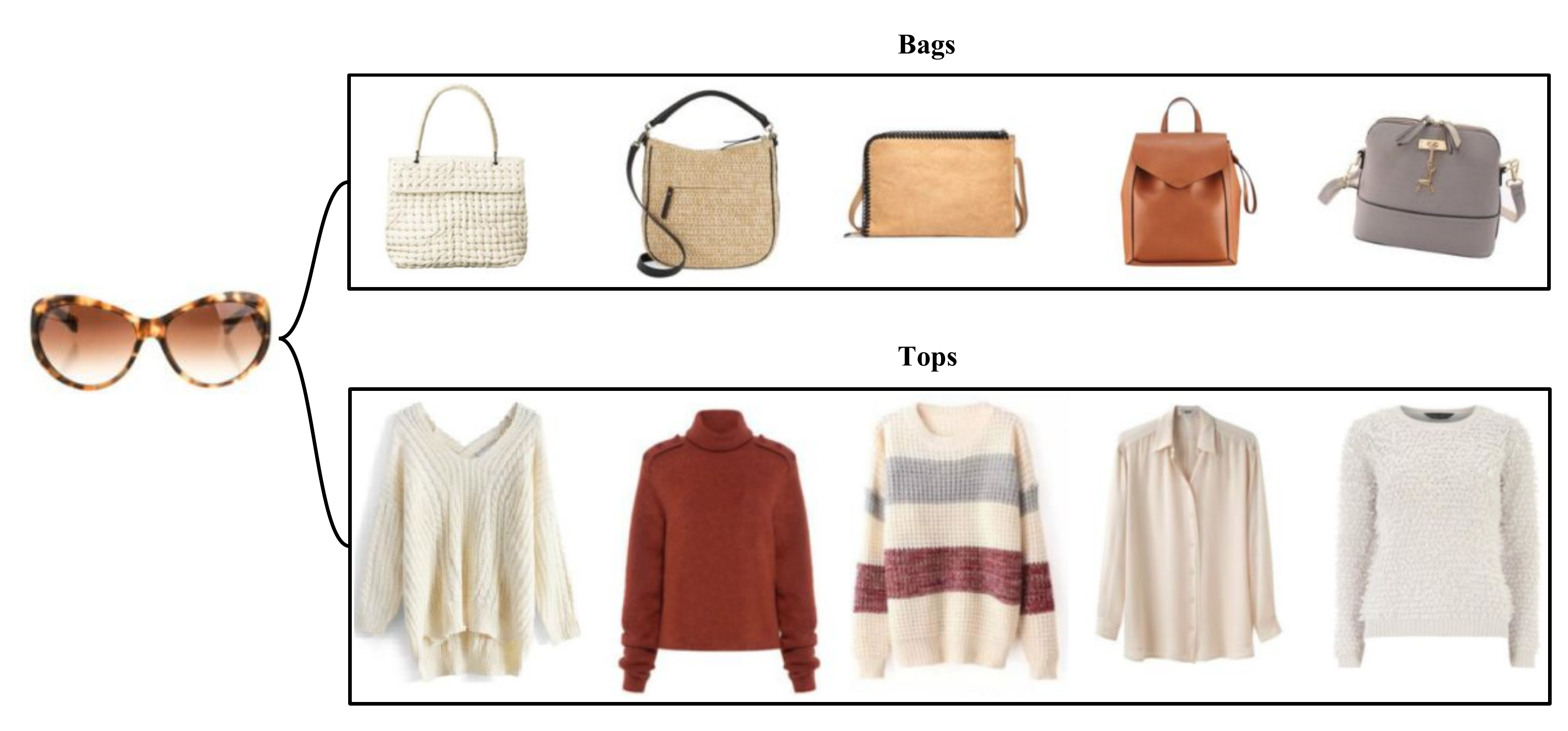}
\end{center}
\caption{The query item is a pair of sunglasses and the items contained in the boxes are the top 5 most compatible bags and tops.}
\label{fig:bags-sunglasses-tops-2}
\end{figure*}

To gain insights into the compatibility relationships learned by our model, we provide visualizations of fashion items belonging to different categories from the Polyvore-Outfits \cite{vasileva2018learning} dataset.  In the following figures, the query item is displayed on the left and items shown in the boxes represent the top 5 most compatible items, as determined by our model, of the specified categories.  We note that the labels specified at the top of the boxes simply denote the categories of the items contained within the boxes and \emph{do not} indicate that they are inputs to our model.  As observed from the figures, color is usually a dominating factor in modeling fashion compatibility.  Fashion items are often deemed to be compatible to each other if they are of similar or complementary shades of color (\eg blue and white in Figure \ref{fig:top-bottom-shoes-2}, and brown and beige in Figure \ref{fig:bags-sunglasses-tops-2}).  Besides the dominating factor of color in the fashion domain, stylistic representations are also shown to be essential in modeling outfit compatibility.  For instance, we observe in Figure \ref{fig:top-bottom-shoes-3} that the specified query shoe item is most compatible with tops and bottoms of flowery designs.


\begin{figure*}[!ht]
\begin{center}
\includegraphics[width=\linewidth, height=8cm]{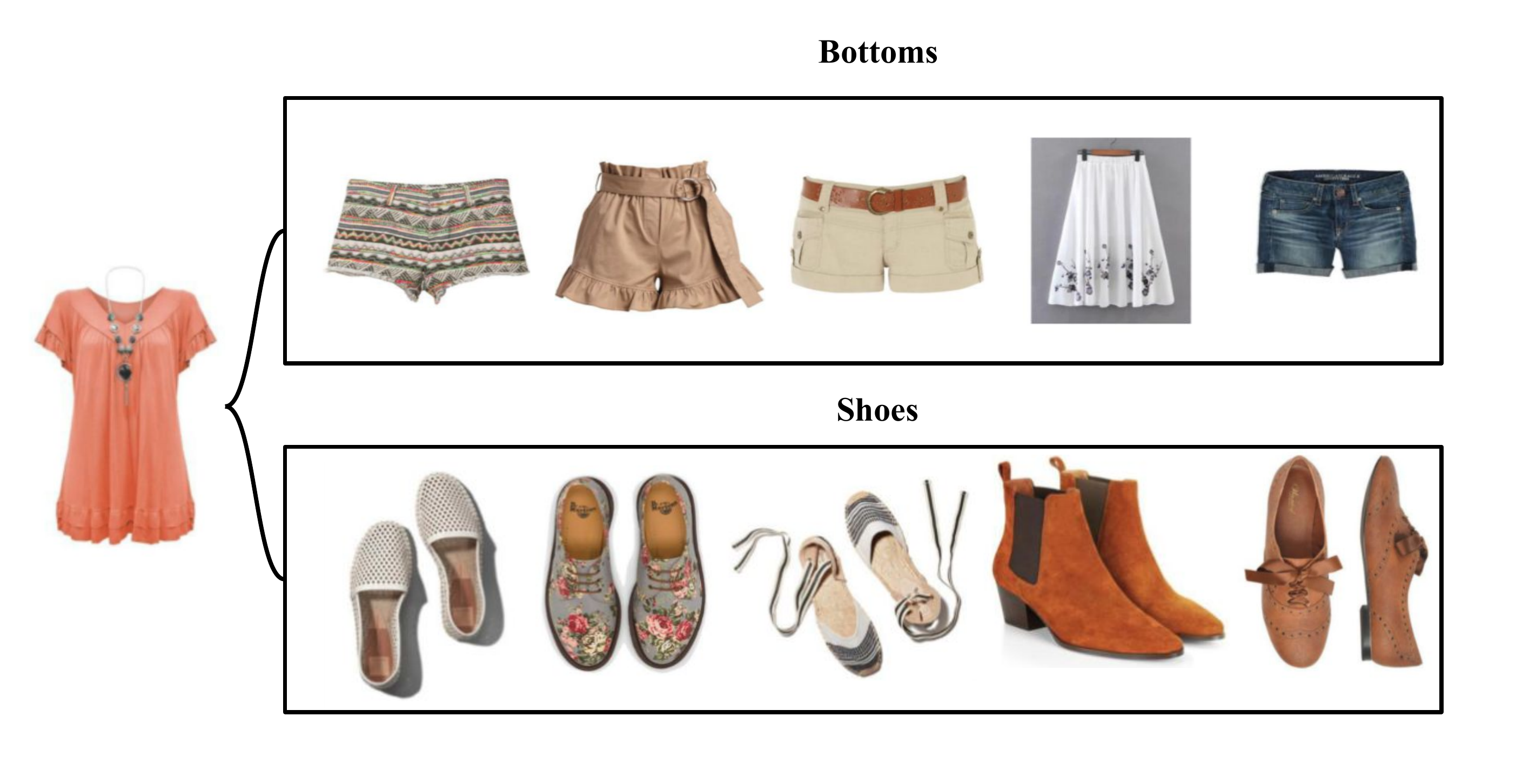}
\end{center}
\vspace{-6mm}
\caption{The query item is a top and the items contained in the boxes are the top 5 most compatible bottoms and shoes.}
\label{fig:top-bottom-shoes-1}
\end{figure*}

\begin{figure*}[!ht]
\begin{center}
\includegraphics[width=\linewidth]{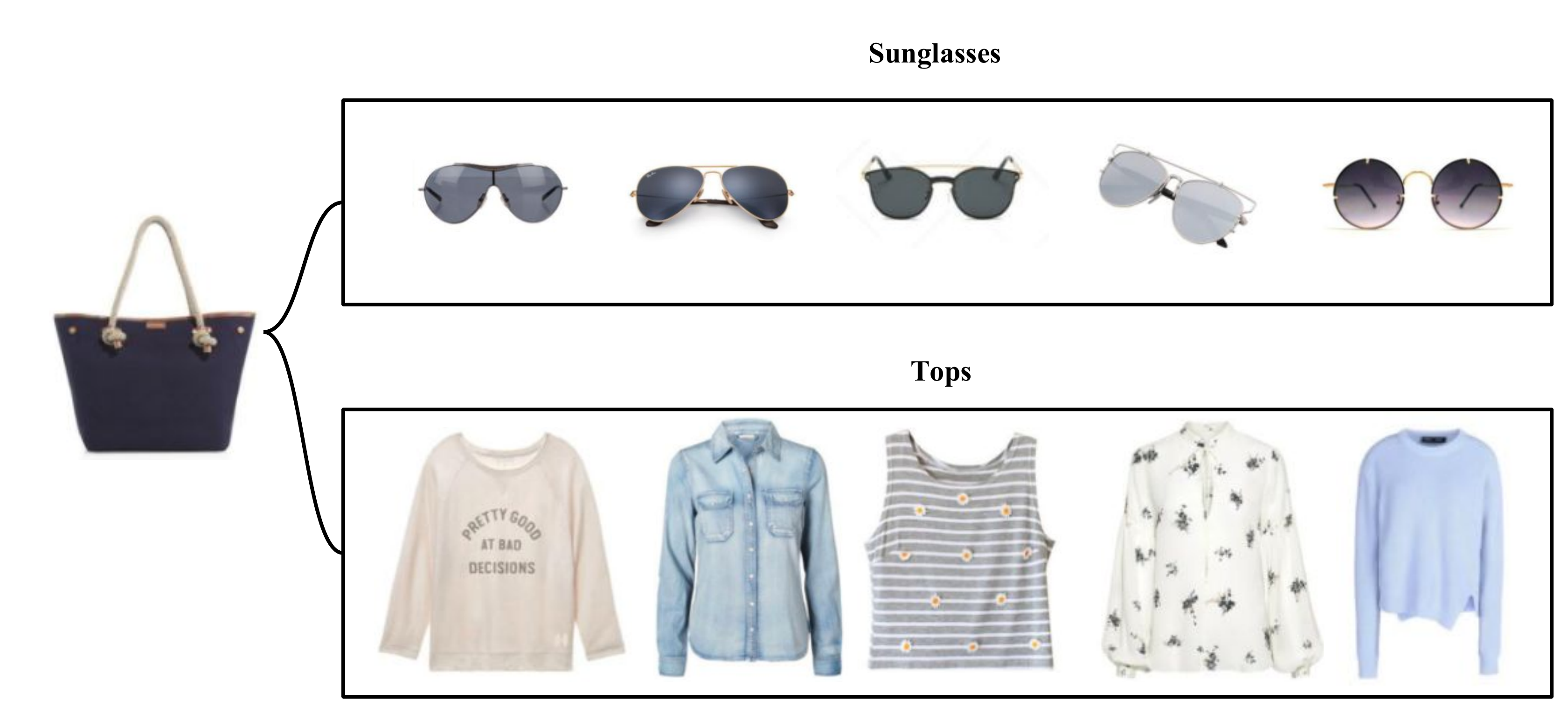}
\end{center}
\caption{The query item is a bag and the items contained in the boxes are the top 5 most compatible sunglasses and tops.}
\label{fig:bags-sunglasses-tops-1}
\end{figure*}

\begin{figure*}[!ht]
\begin{center}
\includegraphics[width=\linewidth]{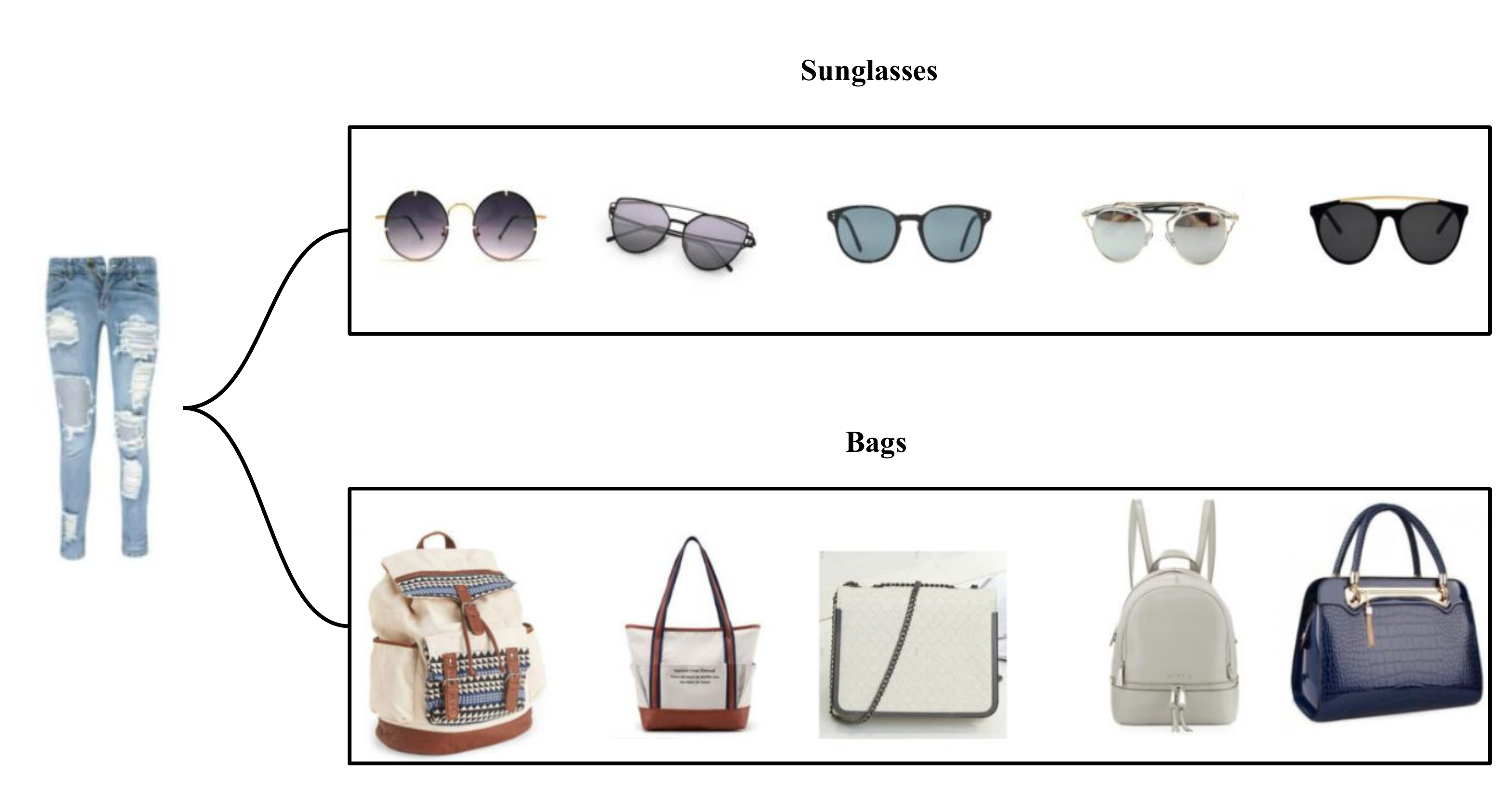}
\end{center}
\caption{The query item is a bottom and the items contained in the boxes are the top 5 most compatible sunglasses and bags.}
\label{fig:bags-bottoms-sunglasses}
\end{figure*}

\begin{figure*}[!ht]
\begin{center}
\includegraphics[width=\linewidth]{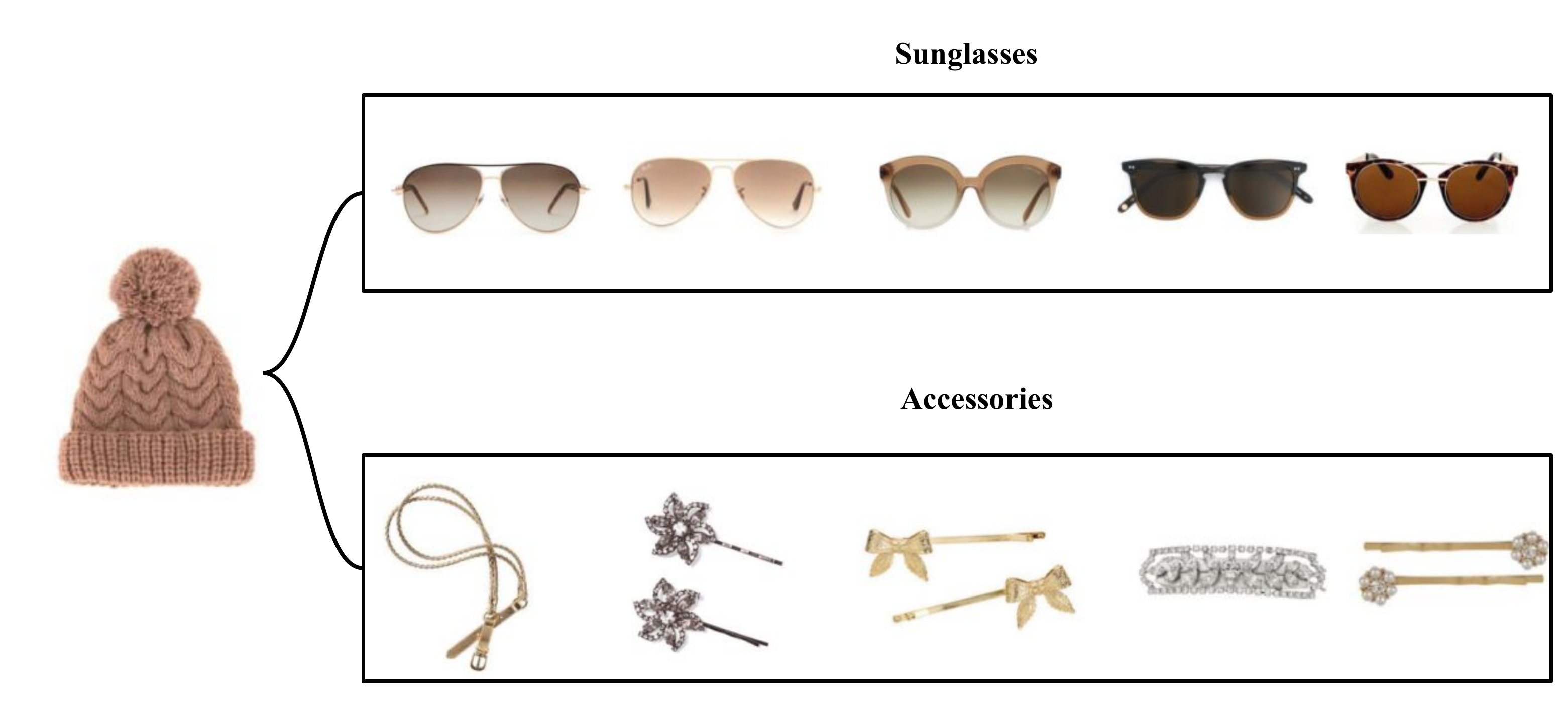}
\end{center}
\caption{The query item is a hat and the items contained in the boxes are the top 5 most compatible sunglasses and earrings.}
\label{fig:acc-hats-sunglasses}
\end{figure*}

\begin{figure*}[!ht]
\begin{center}
\includegraphics[width=\linewidth]{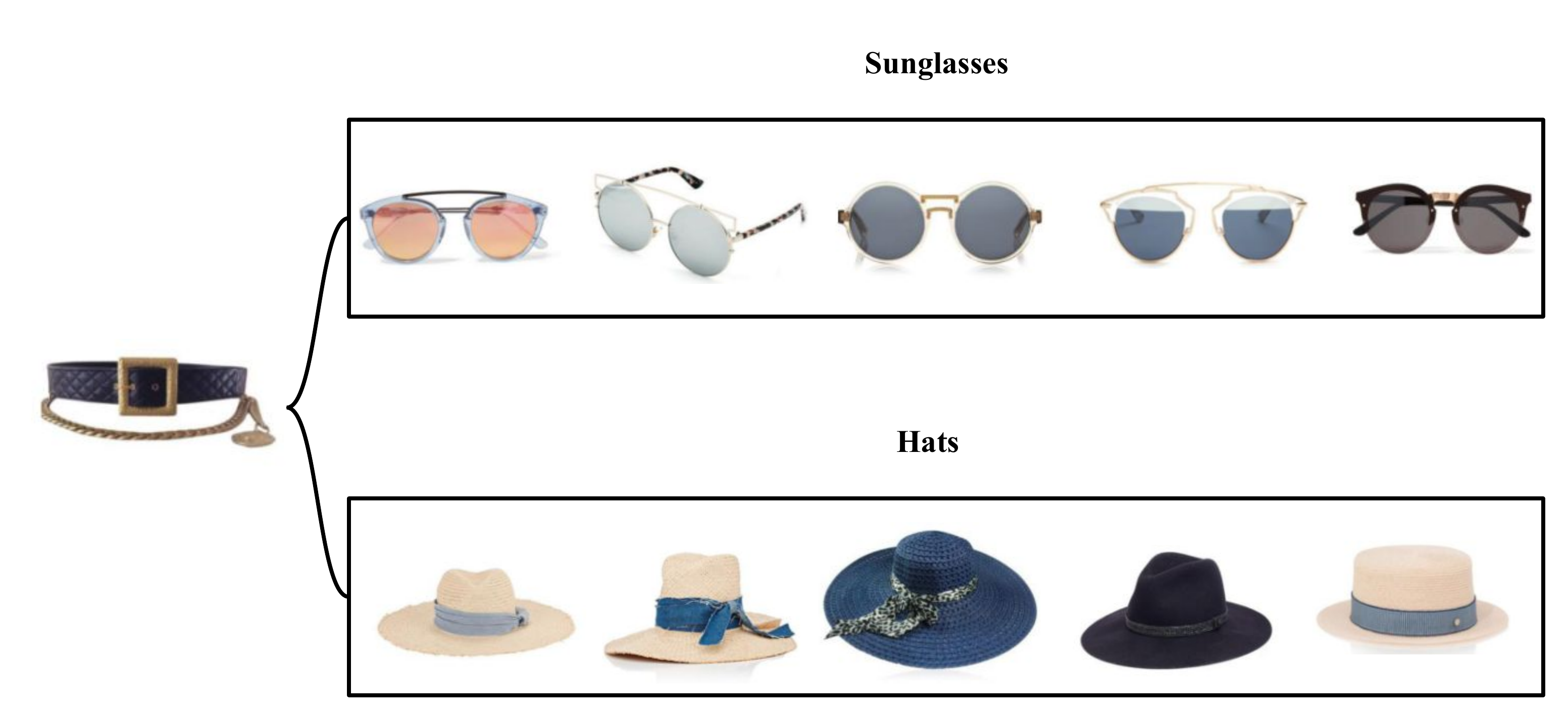}
\end{center}
\caption{The query item is an accessory and the items contained in the boxes are the top 5 most compatible sunglasses and hats.}
\label{fig:acc-hats-sunglasses-2}
\end{figure*}


\begin{figure*}[!ht]
\begin{center}
\includegraphics[width=\linewidth]{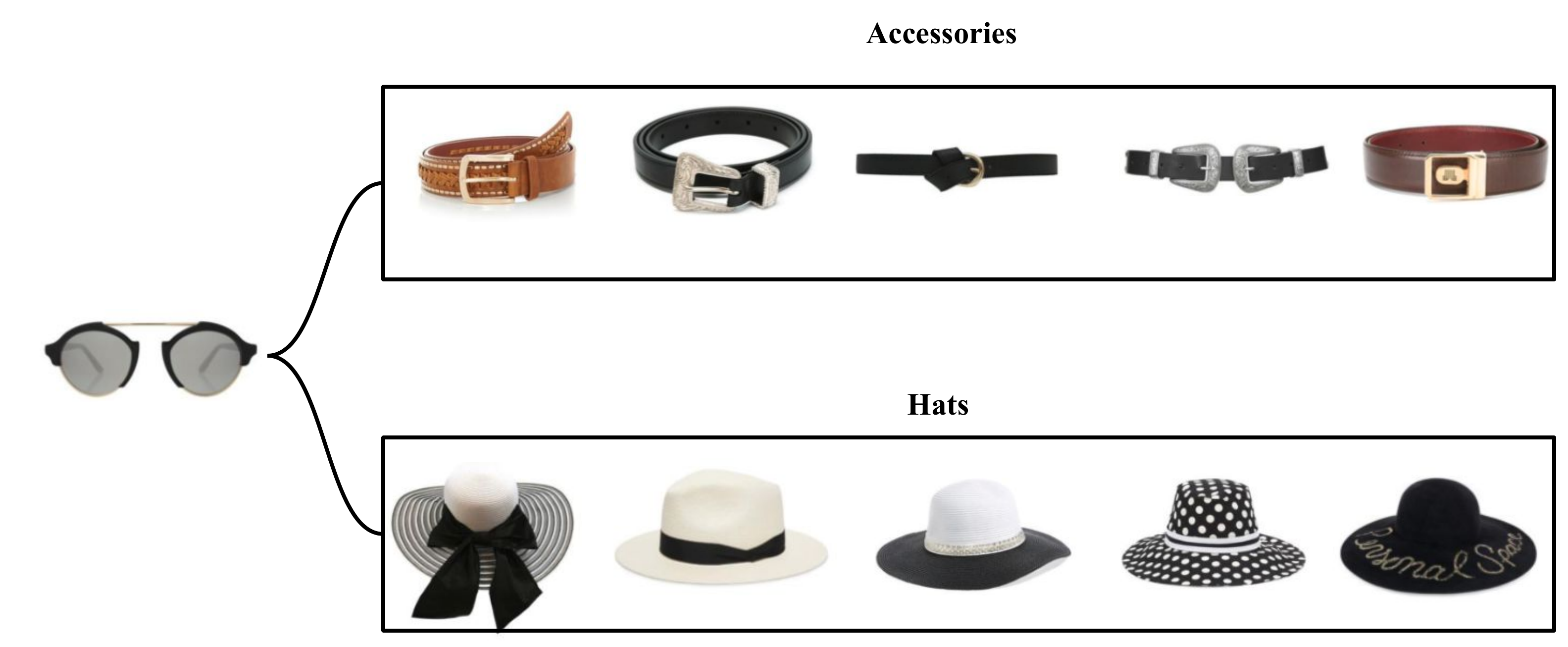}
\end{center}
\caption{The query item is a pair of sunglasses and the items contained in the boxes are the top 5 most compatible belts and hats.}
\label{fig:acc-hats-sunglasses-3}
\end{figure*}




